\documentclass[conference]{IEEEtran}
\IEEEoverridecommandlockouts

\usepackage{amsmath,amssymb,amsfonts}
\usepackage{algorithmic}
\usepackage{graphicx}
\usepackage{textcomp}
\usepackage{xcolor}
\usepackage{pifont}
\usepackage{multirow}
\usepackage[acronym]{glossaries}
\usepackage{siunitx}
\def\BibTeX{{\rm B\kern-.05em{\sc i\kern-.025em b}\kern-.08em
    T\kern-.1667em\lower.7ex\hbox{E}\kern-.125emX}}
\usepackage[colorlinks=true,linkcolor=blue,citecolor=blue]{hyperref}
\usepackage[style=numeric,sorting=none]{biblatex}
\newacronym{FMCW}{FMCW}{Frequency-Modulated Continuous-Wave}
\newacronym{RAMap}{RAMap}{range–azimuth map}
\newacronym{ConfMap}{ConfMap}{confidence map}
\newacronym{RCS}{RCS}{radar cross section}
\newacronym{CFAR}{CFAR}{Constant False Alarm Rate}
\newacronym{MIMO}{MIMO}{Multiple-Input-Multiple-Output}
\newacronym{SBR}{SBR}{shoot-and-bounce-ray}

\newacronym{DDPM}{DDPM}{Denoising Diffusion Probabilistic Model}
\newacronym{SDM}{SDM}{Semantic Diffusion Model}
\newacronym{GAN}{GAN}{Generative Adversarial Network}
\newacronym{VAE}{VAE}{Variational Autoencoder}
\newacronym{CNN}{CNN}{Convolutional Neural Network}
\newacronym{MSE}{MSE}{mean-squared-error}
\newacronym{GAC}{GAC}{geometry-aware conditioning}
\newacronym{TCR}{TCR}{target-consistency regularization}

\newacronym{PSNR}{PSNR}{Peak Signal-to-Noise Ratio}
\newacronym{mAP}{mAP}{mean Average Precision}
\newacronym{AP}{AP}{average precision}
\newacronym{IoU}{IoU}{Intersection-over-Union}
\newacronym{OLS}{OLS}{Object Location Similarity}
\newacronym{NMS}{NMS}{Non-maximum Suppression}

\begin{document}

\title{Synthetic FMCW Radar Range–Azimuth Maps Augmentation with Generative Diffusion Model}

\author{
\IEEEauthorblockN{
Zhaoze Wang\textsuperscript{1, 3},
Changxu Zhang\textsuperscript{1, 3},
Tai Fei\textsuperscript{2},
Christopher Grimm\textsuperscript{1},
Yi Jin\textsuperscript{1}, \\
Claas Tebruegge\textsuperscript{1},
Ernst Warsitz\textsuperscript{1},
Markus Gardill\textsuperscript{3}
}
\IEEEauthorblockA{
\textsuperscript{1}\textit{HELLA GmbH \& Co. KGaA}, Lippstadt, Germany \\
\textsuperscript{2}\textit{Dortmund University of Applied Sciences and Arts}, Dortmund, Germany \\
\textsuperscript{3}\textit{Brandenburg University of Technology}, Cottbus, Germany \\
Email: \{zhaoze.wang, changxu.zhang, christopher.grimm, yi.jin, claas.tebruegge, ernst.warsitz\}@forvia.com, \\
tai.fei@fh-dortmund.de, markus.gardill@b-tu.de}
}

\maketitle

\begin{abstract}
The scarcity and low diversity of well-annotated automotive radar datasets often limit the performance of deep-learning-based environmental perception. To overcome these challenges, we propose a conditional generative framework for synthesizing realistic \gls{FMCW} radar \glspl{RAMap}. Our approach leverages a generative diffusion model to generate radar data for multiple object categories, including pedestrians, cars, and cyclists. Specifically, conditioning is achieved via \glspl{ConfMap}, where each channel represents a semantic class and encodes Gaussian-distributed annotations at target locations. To address radar-specific characteristics, we incorporate \gls{GAC} and \gls{TCR} into the generative process. Experiments on the ROD2021 dataset demonstrate that signal reconstruction quality improves by \SI{3.6}{dB} in \gls{PSNR} over baseline methods, while training with a combination of real and synthetic
datasets improves overall \gls{mAP} by 4.15\% compared with conventional image-processing-based augmentation. These results indicate that our generative framework not only produces physically plausible and diverse radar spectrum but also substantially improves model generalization in downstream tasks.
\end{abstract}


\section{\textbf{Introduction}}

\IEEEPARstart{R}{adar} is a crucial component in autonomous driving, providing robust environmental perception in adverse conditions such as fog, rain, and darkness, where other sensors like cameras and LiDAR often degrade. 

Conventional radar object detection applies the Discrete Fourier Transform (DFT) to convert raw Analog-to-Digital Converter (ADC) samples into Range–Doppler–Azimuth representations, followed by \gls{CFAR} based algorithms. However, the performance of this approach deteriorates considerably under practical conditions due to multi-path reflections, clutter, and limited angular resolution. Although recent advances in deep learning have shown promise in addressing these challenges, the development of radar perception remains constrained by the lack of large-scale, high-quality annotated datasets. While multiple radar datasets have been released in recent years, e.g., \cite{raddet, cruw, kradar}, they are still substantially smaller and less comprehensively labeled than camera- or LiDAR-based benchmarks such as KITTI \cite{kitti} and nuScenes \cite{nuscenes}.

Two common strategies have been explored to address the scarcity of annotated radar data. The first is physical-based radar simulation, which models electromagnetic propagation through ray tracing \cite{rt_Schasler2021, rt_Zong_2023}. These simulators provide accurate ground truth but are computationally expensive, and the artifacts in the real radar measurements are insufficiently modeled or even neglected. Another approach leverages cross-modal supervision, where radar models are trained using pseudo-labels derived from other sensor modalities \cite{Jin2023, radgps}. Although effective in reducing annotation cost, this method depends on precise alignments, inherits biases from teacher modalities and tends not to take full advantage of radar perception.

\begin{figure}[t]
\includegraphics[width=0.48\textwidth]{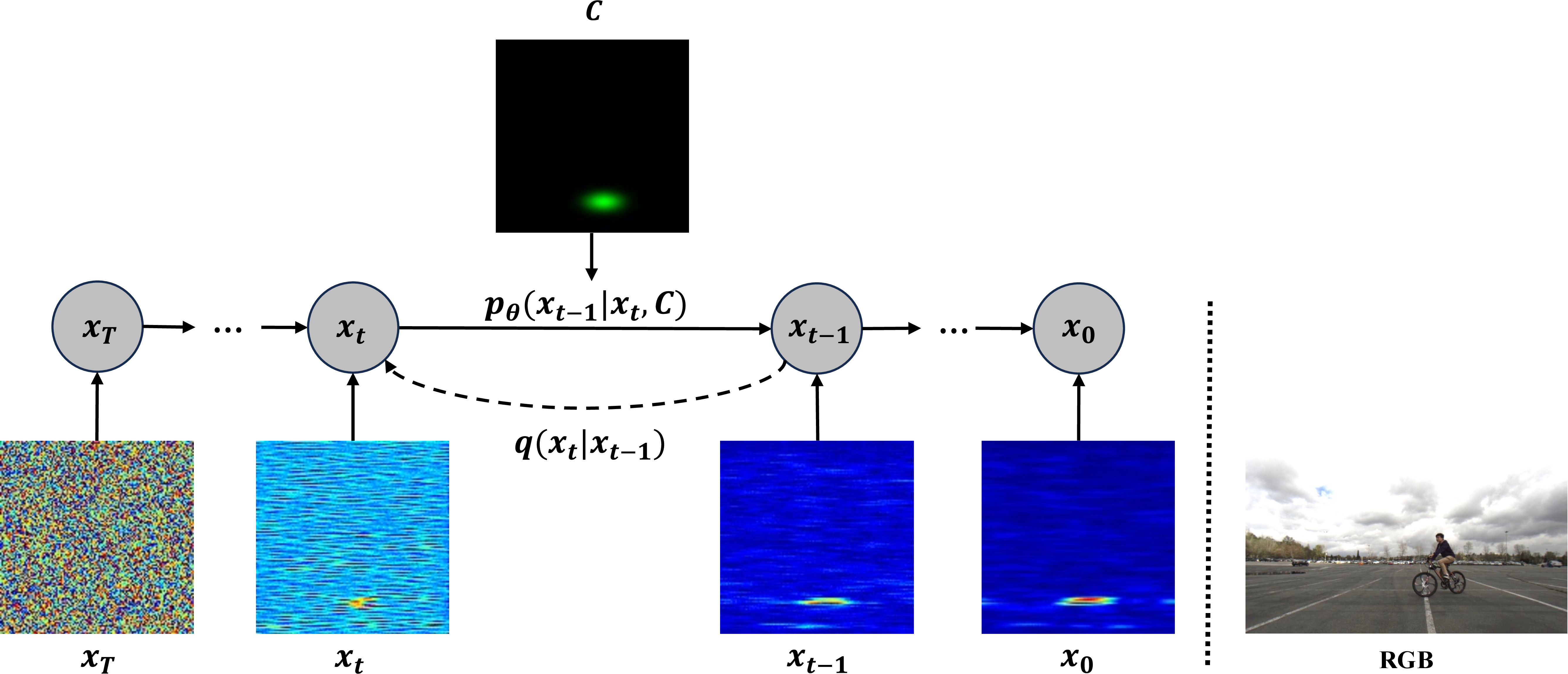}
\caption{Diffusion Model for \glspl{RAMap} reconstruction conditioned by \glspl{ConfMap}. A U-Net-based network $\mathbf{p_\theta(\cdot)}$ predicts noise involved in the noisy image $\mathbf{x_t}$ under the guidance of ConfMap $\mathbf{C}$ obtained from the approach detailed in Sec.\ref{methodology}, and timestamp $t$ in each denoising step to reconstruct the realistic image $\mathbf{x_0}$ iteratively. The RGB image is only used for scene visualization.}
\label{fig1: introduction}
\end{figure}

Most recently, generative models are emerging as a promising paradigm for scalable and realistic data synthesis in autonomous driving. Frameworks such as Pix2Pix \cite{pix2pix} and \gls{DDPM} \cite{ddpm} have shown remarkable performance in generating high-fidelity and diverse data for image modalities \cite{Layout-to-Image}, largely mitigating the dependence on manual annotation. However, the adaptation of generative modeling to the radar domain remains limited, as radar data differ fundamentally from visual data in their non-normalized amplitudes, multi-path propagation, attenuation, and stochastic noise patterns.

In this work, we introduce a conditional generative framework based on the \gls{SDM} \cite{sdm} to synthesize \gls{FMCW} radar \glspl{RAMap}, specifically tailored for downstream radar perception tasks, as shown in Fig.~\ref{fig1: introduction}. Our main contributions are threefold: (1) We introduce a conditional diffusion framework guided by \glspl{ConfMap} for generating realistic \gls{FMCW} radar \glspl{RAMap}. (2) This framework incorporates radar-specific adaptations for the characteristics of radar spectrum, including \gls{GAC} and \gls{TCR}. (3) We comprehensively evaluate the quality of generated signals in terms of both signal-level fidelity and the impact on network performance in the downstream task.

\section{\textbf{Related Works}}

\subsection{Radar Data Generation}

Researchers primarily relied on physics-based simulation as the main approach for generating synthetic radar signal. These simulators model electromagnetic wave propagation using ray tracing techniques to reproduce realistic radar returns in complex driving environments. For instance, a realistic \gls{MIMO} radar simulator based on the \gls{SBR} method \cite{rt_Schasler2021} can capture urban clutter and antenna array effects, while GPU-accelerated ray launching frameworks \cite{rt_Zong_2023} can achieve near real-time performance. However, the computational cost of such simulations increases dramatically with scene complexity, making them impractical for large-scale dataset augmentation.

In contrast, data-driven approaches leverage generative models to synthesize radar data directly from realistic measurements or other modalities. L2RDaS \cite{L2RDaS} synthesizes spatially informative 4D radar tensors from LiDAR data in existing autonomous driving datasets, while 4DR-P2T \cite{4DRP2T} employs a conditional \gls{GAN} to translate radar point clouds into 4D radar tensors. The work most closely related to ours is \cite{bbx2rdmap}, which generates range–doppler maps from bounding box annotations using \glspl{GAN}. However, it focuses on a single object category and ignores specific features of radar signals such as range-dependent attenuation. In contrast, our method generates multi-class \glspl{RAMap} conditioned on ConfMaps that encode object semantics and spatial attributes, enabling the synthesis of diverse and physically plausible radar spectrum. 

\subsection{Generative Image-to-Image Synthesis}

Image-to-image (I2I) synthesis focuses on generating an output image conditioned on an input image. Early approaches extended \glspl{VAE} \cite{vae} to learn latent representation that captures the input-output mapping. \gls{GAN}-based methods, such as pix2pix \cite{pix2pix}, introduced adversarial training to improve visual realism for paired image translation. Unpaired translation methods, such as CycleGAN \cite{cyclegan}, enable style transfer between domains without paired data. However, \gls{GAN}-based models often suffer from training instability and mode collapse. 

Recently, diffusion models have achieved state-of-the-art image synthesis performance by progressively denoising random noise into coherent outputs. ControlNet \cite{controlnet} incorporates spatially guided normalization and conditional feature modulation to control the denoising process according to input semantic maps or other conditioning signals. The \gls{SDM} \cite{sdm} combines semantic conditioning with iterative denoising to produce label-consistent and spatially coherent images.

Nevertheless, recent studies \cite{dm_reg, ldm_reg} reveal that the \gls{MSE} objective in diffusion models may lead to overly smoothed results since it penalizes pixel-wise deviations. This limitation is particularly evident in radar data, where useful information is sparse in the spectral domain. Therefore, we introduce a \gls{TCR} that encourages the model to focus on the target region while allowing background noise to remain diverse. 


\subsection{Radar Object Detection}

Radar object detection was conventionally based on signal processing techniques such as \gls{CFAR}, which carry out detection across range, Doppler, and angle domains using handcrafted thresholds. Although efficient and interpretable, these methods struggle in cluttered or dynamic environments where multipath reflections and noise dominate. In contrast, recent advances leverage deep learning to learn discriminative features directly from radar representations. RADDet \cite{raddet} introduced an anchor-based detection framework on Range--Azimuth--Doppler tensors, while RAMP-CNN \cite{rampcnn} proposed a multiple-perspective \gls{CNN} that processes range--velocity--angle heatmap sequences. TransRAD \cite{transrad} employed retentive self-attention mechanisms to better align with radar spatial priors, achieving precise 3D detection with reduced computational cost. RODNet \cite{rodnet} utilizes camera–radar fusion to obtain labels and performs object detection directly on \glspl{RAMap}. Therefore, we train the same RODNet models on both the original dataset and a hybrid dataset mixed by real and synthetic RAMaps to evaluate the impact of using augmented \glspl{RAMap} on detection performance, as detailed in the section \ref{sec:task-level-evaluation}.

\section{\textbf{Methodology}}
\label{methodology}

\subsection{Confidence Map Generation}
\label{sec: Confidence Map Generation}

To condition the diffusion generator, we construct a ConfMap $\mathbf{C} \in \mathbb{R}^{N_c\times N_r\times N_\theta}$ pixel-wise aligned to the \gls{RAMap} for each frame, where $N_c$, $N_r$, and $N_\theta$ denote the number of categories, range bins, and azimuth bins, respectively. Each frame contains a list of object annotations $\mathcal{L} = \{(r_k, \theta_k, c_k)\}_{k=1}^N$, where each tuple denotes the range, azimuth, and category of the $k$-th detected object. 
Given the \gls{RAMap} discretization $N_r=N_\theta=128$ bins, the maximum detectable range $r_{\max}$ and azimuth angle $\theta_{\max}$ of radar determined by the waveform parameters, the corresponding coordinates $(i_k, j_k)$ of object $k$ are then mapped to the \gls{RAMap} domain by locating the nearest discrete bin indices along the range and azimuth axes.

Motivated by prior evidence that Gaussian-based ConfMaps have been shown effective for \gls{RAMap}-based object detection~\cite{rodnet}, and unlike the fixed-size binary bounding boxes used in previous work by de Oliveira \textit{et al.} \cite{bbx2rdmap}, we represent the spatial energy distributions of object $k$ on ConfMap $\mathbf{C_k}$ as a smooth, variable 2D Gaussian: 
\begin{equation}
    {C}_k(i,j) = \exp\!\Bigg(
        -\frac{(i-i_k)^2}{2\sigma_{r,k}^2} 
        -\frac{(j-j_k)^2}{2\sigma_{\theta,k}^2}
    \Bigg),
\end{equation}
where $\sigma_{r,k}$ and $\sigma_{\theta,k}$ denote the spatial extent of the object along the range and azimuth directions. Both parameters can be derived from the preset object's physical size $L^{(r)}_{c_k}$ and $L^{(\theta)}_{c_k}$ in the range and azimuth directions and distance $r_k$ via:
\begin{equation}
    \sigma_{r,k} = \frac{L^{(r)}_{c_k}}{\Delta r}, \qquad
    \sigma_{\theta,k} = \arctan\!\left(\frac{L^{(\theta)}_{c_k}}{2 r_k}\right).
\end{equation}

\subsection[GAC]{Geometry-Aware Conditioning} 
\label{sec: gac}

The initial ConfMap only represents the ideal location and energy distribution of objects, while the actual radar signal amplitude should further account for radar characteristics and scene geometry. In addition to the free-space path loss caused by distance and the angle-dependent antenna gain, inter-object occlusion should also be considered.  
Specifically, the corrected peak amplitude $\tilde{C}_k(i,j)$ of the $k_{th}$ object’s Gaussian is modulated by several geometry-aware factors as:

\begin{equation}
    \tilde{C}_k(i,j) = A_r(r_k)\cdot A_\theta(\theta_k)\cdot A_{\text{occ},k} \cdot {C}_k(i,j),
\end{equation}
where $A_r$ denotes the distance attenuation, $A_\theta$ represents the antenna gain term, and $A_{\text{occ},k}$ accounts for attenuation due to occlusion by nearer objects.  

According to the radar equation~\cite{radarequation} the received Amplitude is inversely proportional to the second power of distance:
\begin{equation}
     A_r(r_k) \propto \frac{1}{r_k^2}.
\end{equation}

This relationship reflects the free-space attenuation caused by electromagnetic wave propagation between the radar and the targets.  

Because radar antennas are typically anisotropic, the reflected signal strength also varies with the angle relative to the antenna array.  
Therefore, the angular-related gain term $A_\theta(\theta_k)$ is supposed to be introduced and can be derived from the antenna pattern provided in the Texas Instruments AWR1843BOOST radar datasheet, which was used for data collection in the ROD2021 dataset.  

Occlusion attenuation further considers near–far interactions among objects within close azimuths.  
For a given object $q$, we search for nearer objects $p$ whose azimuth difference satisfies $|\theta_p - \theta_q| < \Delta\theta_{\text{occ}}$, where $\Delta\theta_{\text{occ}}$ is a predefined angular window.  
If such occluders exist, we apply an attenuation coefficient $A_{\text{occ},k}$ (e.g., pedestrians may cause weaker occlusion than cars).  
This simple yet effective modeling helps mimic the visibility variations commonly observed in complex scenes.

The class-wise ConfMap $\mathbf{C^{(i)}}$ is obtained by summing all instance-level $\mathbf {\tilde{C_k}}$ up within each category. Subsequently, the final multi-channel ConfMap is concatenated along the channel dimension as: 
\begin{equation}
    \mathbf{C} = \mathrm{Concat}\left(\mathbf{{C}^{(1)}}, \mathbf{{C}^{(2)}}, \ldots, \mathbf{{C}^{(N_c)}}\right)
    \in \mathbb{R}^{N_c \times N_r \times N_\theta},
\end{equation}
which serves as the conditional input to the generative model.

These geometry-aware corrections explicitly incorporate the physical characteristics of radar signal propagation into the conditioning input, thereby alleviating the burden on the neural network to implicitly learn such latent relations from data.  Some samples of adjusted ConfMaps have been shown in the second row of Fig.~\ref{figures/results_generated_ramap}.

\subsection[TCR]{Target-Consistency Regularization}
\label{sec: tcr}

We formulate radar \gls{RAMap} generation as a conditional image-to-image diffusion process.  
The denoising network $\epsilon_\theta(\mathbf{x_t}, t, \mathbf{C})$ that we introduced from \cite{sdm} predicts the noise added at timestep $t$ given a noisy sample $\mathbf{x_t}$ and a semantic ConfMap $\mathbf{C}$.
The clean estimate $\mathbf{\hat{x}_0}$ can be reconstructed as
\begin{equation}
\mathbf{\hat{x}_0} = \frac{\mathbf{x_t }- \sqrt{1 - \bar{\alpha}_t}\, \epsilon_\theta(\mathbf{x_t}, t, \mathbf{C})}{\sqrt{\bar{\alpha}_t}},
\end{equation}
where $\bar{\alpha}_t$ is the cumulative product of the noise schedule at timestep $t$. Accordingly, the standard diffusion objective minimizes the \gls{MSE} between the true noise $\epsilon$ and the network prediction:
\begin{equation}
\mathcal{L}_{\text{MSE}} = \| \epsilon - \epsilon_\theta(\mathbf{x_t}, t, \mathbf{C}) \|_2^2.
\end{equation}


While \gls{MSE} is theoretically grounded and widely adopted, recent studies have shown that such assumptions cause diffusion models to produce over-smoothed or physically implausible samples since \gls{MSE} assumes pixel-wise independence and penalizes all deviations uniformly \cite{dm_reg, ldm_reg}. A perceptual regularization helps the model concentrate on semantic consistency, but it is computed with backbones pretrained on visual images like VGG, which are unsuitable for radar spectra. Hence, we introduce a differentiable regularization item, inspired by the \gls{CFAR} and formulated using a focal-style objective. The goal is to maximize consistency of the foreground and background probabilistic map between the reconstructed and ground-truth \glspl{RAMap}, while allowing stochastic noise diversity in the background.  

For each pair of reconstructed and ground-truth \gls{RAMap} $\{\mathbf{\hat{x}_0}, \mathbf{x_0}\}$, we compute a spatially adaptive detection threshold $\tau(i, j)$ at pixel location $(i,j)$ using local statistics:
\begin{equation}
\tau(i,j) = \mu(i,j) + w\cdot\sigma(i,j),
\end{equation} where $\mu(i,j)$ and $\sigma(i,j)$ are the local mean and standard deviation computed by average pooling centered at $(i,j)$, and $w$ is a scalar that controls detection sensitivity. This formulation ensures that each region adapts its noise floor dynamically, maintaining a roughly constant false alarm rate across varying clutter levels.

Given the adaptive threshold $\mathbf{\tau}$, we can compute per-pixel probability maps for the reconstructed and ground-truth \glspl{RAMap} as
\begin{equation}
\mathbf{p_{\hat{x}_0}} = f_\sigma\!(\alpha (\mathbf{\hat{x}_0} - \mathbf{\tau})), \qquad
\mathbf{p_{x_0}} = {f_\sigma\!} (\alpha(\mathbf{x_0} - \mathbf{\tau})),
\end{equation}
where $f_\sigma(\cdot)$ denotes the sigmoid function, and a scalar $\alpha$ controls the sharpness of the transition between foreground and background.

Instead of a simple $\ell_1$ or $\ell_2$ distance between probability maps, we adopt a focal-loss \cite{focalloss} formulation that prioritizes uncertain target regions:
\begin{equation}
\begin{split}
\mathcal{L}_{\text{TCR}} = - \sum \Big[ & \,
\mathbf{p_{x_0}} \, (1 - \mathbf{p_{\hat{x}_0}})^\gamma \, \log(\mathbf{p_{\hat{x}_0}}) \\
& + (1 - \mathbf{p_{x_0}}) \, (\mathbf{p_{\hat{x}_0}})^\gamma \, \log(1 - \mathbf{p_{\hat{x}_0}}) 
\Big],
\end{split}
\end{equation}

where parameter $\gamma$ is a tunable focusing parameter to emphasize hard discriminable samples, allowing the model to concentrate more on learning challenging categories while suppressing trivial background, which is also well-suited for radar spectrograms prediction, where target signals usually occupy notably limited portions in the spectrum \cite{radgps}.

The final training objective combines the diffusion reconstruction term and the proposed \gls{TCR} term:
\begin{equation}
\mathcal{L}_{\text{total}} = 
\mathcal{L}_{\text{MSE}} + \lambda_{\text{TCR}} \mathcal{L}_{\text{TCR}},
\end{equation}
where $\lambda_{\text{TCR}}$ balances the influence of the regularization. 

\section{\textbf{Experiments}}

\subsection{Dataset}
\label{sec:dataset}

We adopt the ROD2021 dataset released by Y. Wang \textit{et al.} ~\cite{rodnet}. The annotated object categories include \textit{pedestrian}, \textit{cyclist}, and \textit{car}. The dataset contains four distinct scenes: \textit{Parking Lot}, \textit{Campus Road}, \textit{City Street}, and \textit{Highway}, comprising a total of 40 sequences, and was split into 80\% for training and 20\% for testing, while the test set includes samples from all four scenes to ensure comprehensive evaluation. 

The sample statistics for each scene in both the original and hybrid datasets are summarized in Table~\ref{tab:data_distribution_mixed}, where it indicates that the \textit{City Street} and \textit{Highway} scenes contain significantly fewer samples than the average in the original dataset. To mitigate this imbalance, we augment the training set with our synthetic data in these two scenes. Specifically, 3D bounding boxes annotations of 8 sequences are randomly selected from the K-Radar dataset \cite{kradar}, and these 3D labels can be converted into range-azimuth trajectories for preparing ConfMaps following the procedure described in Section~\ref{methodology}. To maintain a consistent total dataset size, an equivalent number of frames is reduced from the other two scenes, forming the final \textit{Hybrid Dataset} for training of the downstream task.

\begin{table}[h!]
\centering
\caption{Statistics of the original and hybrid ROD2021 datasets. Each value denotes the number of sequences or total frames per scene.}
\begin{tabular}{l cc cc}
\hline
\hline
\multirow{2}{*}{\textbf{Scene}} & 
\multicolumn{2}{c}{\textbf{Original Dataset}} & 
\multicolumn{2}{c}{\textbf{Hybrid Dataset}} \\
\cline{2-5}
 & \textbf{\# of Seq} & \textbf{\# of Frames} & \textbf{\# of Seq} & \textbf{\# of Frames} \\
\hline
Parking Lot   & 22 & 19{,}767 & 16 & 14{,}702 \\
Campus Road   & 12 & 10{,}305 & 10 & 8{,}505 \\
City Street   & 2  & 2{,}908  & 6  & 5{,}288 \\
Highway       & 4  & 5{,}105  & 8  & 7{,}493 \\
\hline
Overall       & 40 & 38{,}085 & 40 & 35{,}988 \\
\hline
\hline
\end{tabular}
\label{tab:data_distribution_mixed}
\end{table}

\subsection{Training Details}
\label{sec:training-details}

For the generative task,  we introduce the \gls{SDM} model in this work that follows a U-Net architecture with dedicated attention modules for integrating conditions into the network. For the radar object detection task, we employ the HG1V2-DCN model from the RODNet framework~\cite{rodnet}, which serves as a baseline detector trained under data with different augmentation strategies.

All experiments were conducted on a single NVIDIA RTX 5080 GPU with an Intel i9-14900KF CPU. The model is trained for 50 epochs with a batch size of 4 using the Adam optimizer with an initial learning rate of $3\times10^{-5}$ and standard weight decay to $1\times10^{-8}$.

\subsection{Signal-Level Evaluation}

To quantitatively assess the quality of the generated radar \glspl{RAMap}, we evaluate the signal-level reconstruction using the \gls{PSNR}.  
Given a generated map $\hat{x_0}$ and its ground truth $x_0$, \gls{PSNR} measures their pixel-wise similarity in the logarithmic domain:
\begin{equation}
\mathrm{PSNR} = 10 \log_{10} \left(\frac{A_{\max}^2}{\frac{1}{N}\sum_{i=1}^{N} (x_i - \hat{x}_i)^2}\right),
\end{equation}
where $A_{\max}$ denotes the maximum signal amplitude and $N$ is the number of pixels.  
Higher \gls{PSNR} values indicate better signal fidelity and less reconstruction distortion.

\begin{table}[h]
\caption{Signal-Level Evaluation (\gls{PSNR} in \si{dB}, higher is better). $\checkmark$ denotes the use of each component.}
\begin{center}
\begin{tabular}{c|c|c|c|c}
\hline
\hline
\textbf{Method} & \textbf{Condition} & \nameref{sec: gac} & \nameref{sec: tcr} & \textbf{PSNR$\uparrow$} \\
\hline
de Oliveira \textit{et al.} \cite{bbx2rdmap} & BBX Mask & $\times$ & $\times$ & 20.1 \\
Ours & ConfMap & $\times$ & $\times$ & 22.1 \\
Ours & ConfMap & $\checkmark$ & $\times$ & 23.3 \\
\textbf{Ours} & ConfMap & $\checkmark$ & $\checkmark$ & \textbf{23.7} \\
\hline
\hline
\end{tabular}
\label{tab:signal_psnr}
\end{center}
\end{table}

\begin{figure*}[!t]
\centering
\includegraphics[width=\textwidth]{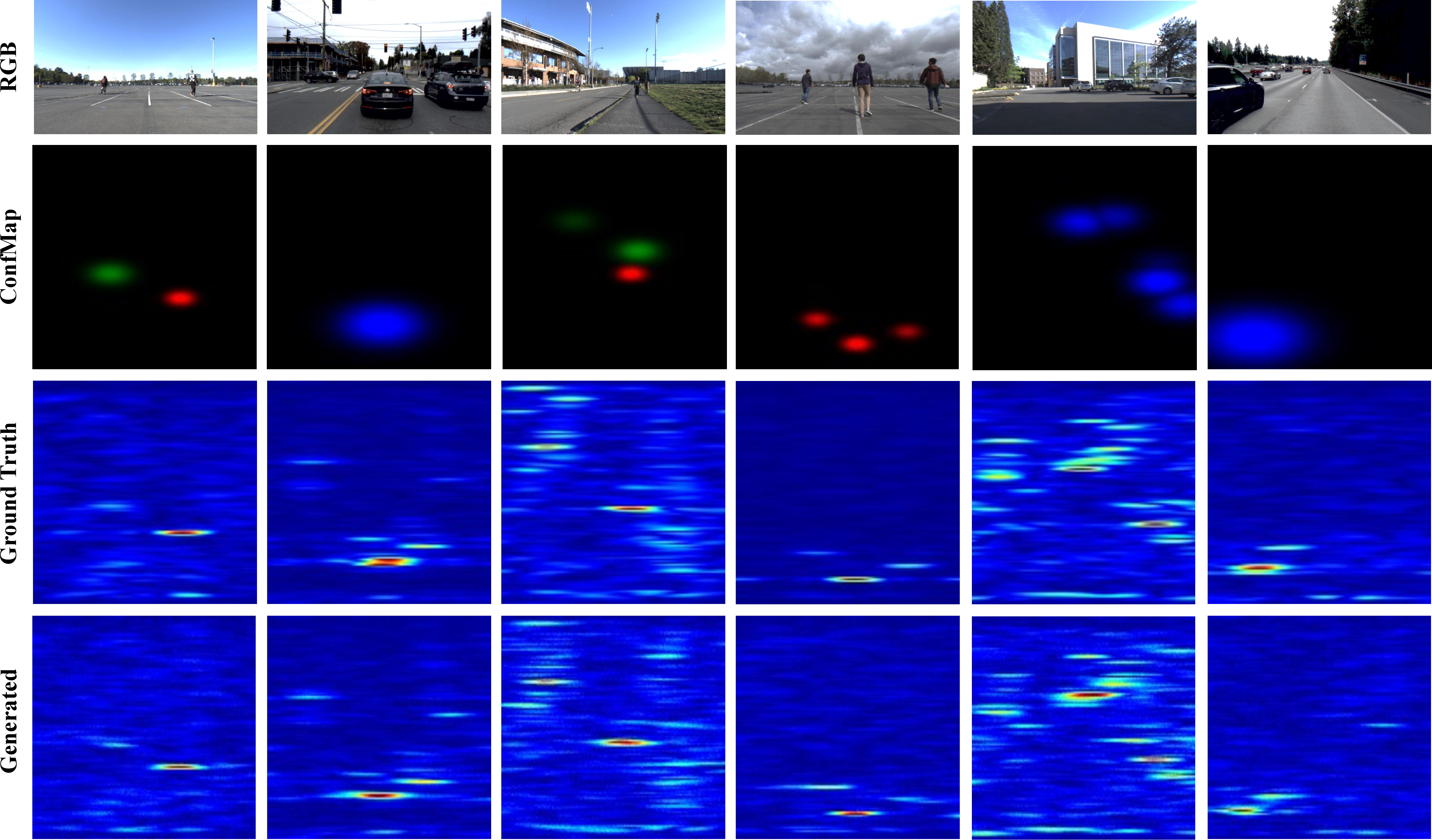}
\caption{Qualitative results of our \gls{RAMap} generation. Each example is organized into four rows. The first row shows the original RGB images. The second row presents the ConfMaps for different object categories, where red indicates pedestrians, green indicates cyclists, and blue indicates cars. The third row shows the ground-truth \gls{RAMap}, and the fourth row displays the \glspl{RAMap} generated by our model. }
\label{figures/results_generated_ramap}
\end{figure*}

\begin{table*}[!t]
\centering
\caption{Radar object detection performance comparison on the ROD2021 dataset and its augmented variants. Metrics include \gls{mAP} and \gls{AP} at an \gls{OLS} threshold of 0.5.}
\label{tab:radar_detection_results}
\resizebox{\textwidth}{!}{
\begin{tabular}{l|cc|cc|cc|cc|cc}
\hline
\hline
\multirow{2}{*}{\textbf{Method}} &
\multicolumn{2}{c|}{\textbf{Parking Lot}} &
\multicolumn{2}{c|}{\textbf{Campus Road}} &
\multicolumn{2}{c|}{\textbf{City Street}} &
\multicolumn{2}{c|}{\textbf{Highway}} &
\multicolumn{2}{c}{\textbf{Overall}} \\ \cline{2-11}
&{mAP}(\%)$\uparrow$ & AP@0.5(\%)$\uparrow$ 
&{mAP}(\%)$\uparrow$ & AP@0.5(\%)$\uparrow$
&{mAP}(\%)$\uparrow$ & AP@0.5(\%)$\uparrow$ 
&{mAP}(\%)$\uparrow$ & AP@0.5(\%)$\uparrow$
&{mAP}(\%)$\uparrow$ & AP@0.5(\%)$\uparrow$ \\ 
\hline
RODNet~\cite{rodnet} & 
\textbf{58.41} & \textbf{59.61} & 
34.27 & 36.89 &
17.71 & 23.59 &
30.24 & 31.32 &
31.30 & 35.22 \\

RAMP-CNN~\cite{rampcnn} & 
57.95 & 59.10 &
35.10 & 36.45 &
18.25 & 23.10 &
30.90 & 31.50 &
31.50 & 35.10 \\

\textbf{Ours} & 
57.35 & 58.42 &
\textbf{36.95} & \textbf{37.15} &
\textbf{21.01} & \textbf{25.37} &
\textbf{33.82} & \textbf{34.07} &
\textbf{35.65} & \textbf{39.75}\\
\hline
\hline
\end{tabular}
}
\end{table*}

We compare our method against the baseline method proposed by de Oliveira \textit{et al.} \cite{bbx2rdmap} that utilizes bounding-box masks as conditional input, and also validate the effect of enabling the proposed \gls{GAC} and \gls{TCR}. Qualitative results are shown in Fig.~\ref {figures/results_generated_ramap} and quantitative results are listed in Table~\ref{tab:signal_psnr}, which illustrate that our method achieves higher \gls{PSNR} values than the baseline method. Moreover, incorporating the proposed geometry- and regularization-based enhancements further improves \gls{PSNR} by up to \SI{1.6}{dB}, demonstrating their effectiveness in enforcing physical consistency and realism of the synthesized radar signals.

\subsection{Task-Level Evaluation}
\label{sec:task-level-evaluation}

To evaluate the fidelity of the generated \glspl{RAMap} and their impact on downstream radar perception, 
we train RODNet~\cite{rodnet} on both the original and hybrid datasets. Detection performance is assessed using the \gls{OLS} metric~\cite{rodnet}, 
which replaces the \gls{IoU} for object detection. \gls{OLS} measures the similarity between a predicted and ground-truth object on the ConfMap as
\begin{equation}
\mathrm{OLS} = \exp\!\left(-\frac{\Delta d^2}{2(r\,\kappa_c)^2}\right),
\end{equation}
where $\Delta d$ denotes the Euclidean distance (in meters) between the predicted and ground-truth object centers, 
$r$ is the distance of the object from the radar sensor, 
and $\kappa_c$ is a class-dependent tolerance factor determined by the typical object size of class $\mathrm{c}$. 
A higher \gls{OLS} value indicates better spatial alignment and scale consistency.

Following~\cite{rodnet}, we compute \gls{AP} at \gls{OLS} thresholds $\tau \in \{0.5, 0.55, \ldots, 0.9\}$, 
denoted as $\mathrm{AP@OLS}_\tau$. 
For a given class $c$ and threshold $\tau$, the AP is defined as
\begin{equation}
\mathrm{AP@OLS}_{c,\tau} = \int_0^1 p_{c,\tau}(r)\,dr,
\end{equation}
where $p_{c,\tau}(r)$ is the precision–recall curve for class $c$ under threshold $\tau$. 
The mean across all classes and thresholds yields the final \gls{mAP}:
\begin{equation}
\mathrm{mAP} = 
\frac{1}{N_c N_\tau} \sum_{c=1}^{N_c} \sum_{\tau \in \mathcal{T}} 
\mathrm{AP@OLS}_{c,\tau}.
\end{equation}
\gls{NMS} is also applied to remove redundant detections and retain only the predictions with the highest score.

For comparison, we also implement the image-processing-based \gls{RAMap} augmentation proposed by X. Gao \textit{et al.} ~\cite{rampcnn}, which performs random translations and flips along range and azimuth directions. As shown in Table~\ref{tab:radar_detection_results}, training on the hybrid dataset does not degrade performance and consistently improves detection accuracy, particularly for underrepresented scenes such as \textit{City Street} and \textit{Highway}. These results indicate that our method can effectively augment radar datasets using existing labeled samples, providing a practical strategy for improving object detection under data-limited conditions.

\section{\textbf{Conclusion}}
This work presents a conditional generative framework to address the scarcity of annotated radar data for deep-learning perception. We synthesize realistic \gls{FMCW} radar \glspl{RAMap} guided by ConfMaps. To adapt the diffusion process to radar-specific characteristics, two strategies are introduced. First, \gls{GAC} models visibility, ensuring overlapping targets yield physically consistent radar responses. Second, a \gls{TCR} term encourages the network to focus on target energy distributions while allowing background variation.

Experiments on the ROD2021 dataset demonstrate the framework's effectiveness. The generated signals exhibit improved realism and physical plausibility, achieving a \SI{3.6}{dB} \gls{PSNR} gain over baseline. Training downstream radar object detectors with the augmented dataset leads to a 4.15\% increase in\gls{PSNR}. These results confirm that the proposed framework produces diverse, semantically consistent radar signatures that enhance radar perception generalization.

Future work will explore more efficient generative paradigms, such as flow matching, to reduce sampling latency while maintaining generation quality. Furthermore, generating range–Doppler features and micro-Doppler Effect may further support finer-grained radar perception tasks.

\section{\textbf{Acknowledgement}} 
The research leading to these results is funded by the German Federal Ministry for Economic Affairs and Energy within the project “NXT GEN AI METHODS – Generative Methoden für Perzeption, Prädiktion und Planung". The authors would like to thank the consortium for the successful cooperation.


\end{document}